\documentclass[letterpaper, 10pt, conference]{ieeeconf}
\IEEEoverridecommandlockouts

\usepackage{cite}
\usepackage{float}
\usepackage{graphicx}
\usepackage{amssymb} 
\usepackage{mathrsfs}
\usepackage{amsmath}
\usepackage{algorithm}
\usepackage[noend]{algpseudocode}
\usepackage{array}
\usepackage{caption}
\usepackage{subcaption}
\usepackage{url}
\usepackage[normalem]{ulem}
\useunder{\uline}{\ul}{}
\usepackage{multirow}

\usepackage{xcolor}
\usepackage[colorlinks]{hyperref} 
\hypersetup{
    colorlinks=true,
    breaklinks=true, 
    urlcolor= blue,              
    linkcolor= blue, 
    bookmarksopen=false, 
    filecolor=black, 
    citecolor=blue, 
    linkbordercolor=blue
}

\usepackage{svg}
\usepackage{booktabs}

\newcommand{\Rv}{\mathbf{R}}

\newcommand{\Tv}{\mathbf{T}}

\title{\LARGE \bf  DytanVO: Joint Refinement of Visual Odometry and Motion Segmentation in Dynamic Environments}

\author{Shihao~Shen, Yilin~Cai, Wenshan~Wang, Sebastian~Scherer
\thanks{Code is available at \href{https://github.com/Geniussh/DytanVO}{https://github.com/Geniussh/DytanVO}}
\thanks{S. Shen, Y. Cai, W. Wang, S. Scherer are with the Robotics Institute, Carnegie Mellon University, Pittsburgh, PA 15213, USA. {\tt\small \{shihaosh, yilincai, wenshanw, basti\}@andrew.cmu.edu}}
}

\begin{document}


\maketitle

\begin{abstract}
    Learning-based visual odometry (VO) algorithms achieve remarkable performance on common static scenes, benefiting from high-capacity models and massive annotated data, but tend to fail in dynamic, populated environments. Semantic segmentation is largely used to discard dynamic associations before estimating camera motions but at the cost of discarding static features and is hard to scale up to unseen categories. In this paper, we leverage the mutual dependence between camera ego-motion and motion segmentation and show that both can be jointly refined in a single learning-based framework. In particular, we present DytanVO, the first supervised learning-based VO method that deals with dynamic environments. It takes two consecutive monocular frames in real-time and predicts camera ego-motion in an iterative fashion. Our method achieves an average improvement of 27.7\% in ATE over state-of-the-art VO solutions in real-world dynamic environments, and even performs competitively among dynamic visual SLAM systems which optimize the trajectory on the backend. Experiments on plentiful unseen environments also demonstrate our method's generalizability. 
\end{abstract}


\IEEEpeerreviewmaketitle

\section{Introduction}


Visual odometry (VO), one of the most essential components for pose estimation in the visual Simultaneous Localization and Mapping (SLAM) system, has attracted significant interest in robotic applications over past few years~\cite{scaramuzza2011visual}. A lot of research works have been conducted to develop an accurate and robust monocular VO system using both geometry-based methods~\cite{engel2017direct,forster2014svo}. However, they require significant engineering effort for each module to be carefully designed and finetuned~\cite{wang2017deepvo}, which makes it difficult to be readily deployed in the open world with complex environmental dynamcis, changes of illumination or inevitable sensor noises.

On the other hand, recent learning-based methods~\cite{tartanvo,wang2017deepvo,DeepTAMDT,li2020self} are able to outperform geometry-based methods in more challenging environments such as large motion, fog or rain effects and lack of features. However, they will easily fail in dynamic environments if they do not take into consideration independently moving objects that cause unpredictable changes in illumination or occlusions. To this end, recent works utilize abundant unlabeled data and adopt either self-supervised learning~\cite{xuan2022maskvo, kim2022simvodis++} or unsupervised learning~\cite{yin2018geonet, ranjan2019competitive} to handle dynamic scenes. Although they achieve outstanding performance on particular tasks, such as autonomous driving, they produce worse results if applied to very different data distributions, such as micro air vehicles (MAV) that operate with aggressive and frequent rotations cars do not have. Learning without supervision is hindered from generalizing due to biased data with simple motion patterns. Therefore, we approach the dynamic VO problem as supervised learning so that the model can map inputs to complex ego-motion ground truth and be more generalizable.
\begin{figure}[t!]
    \centering
    \includegraphics[width = \columnwidth]{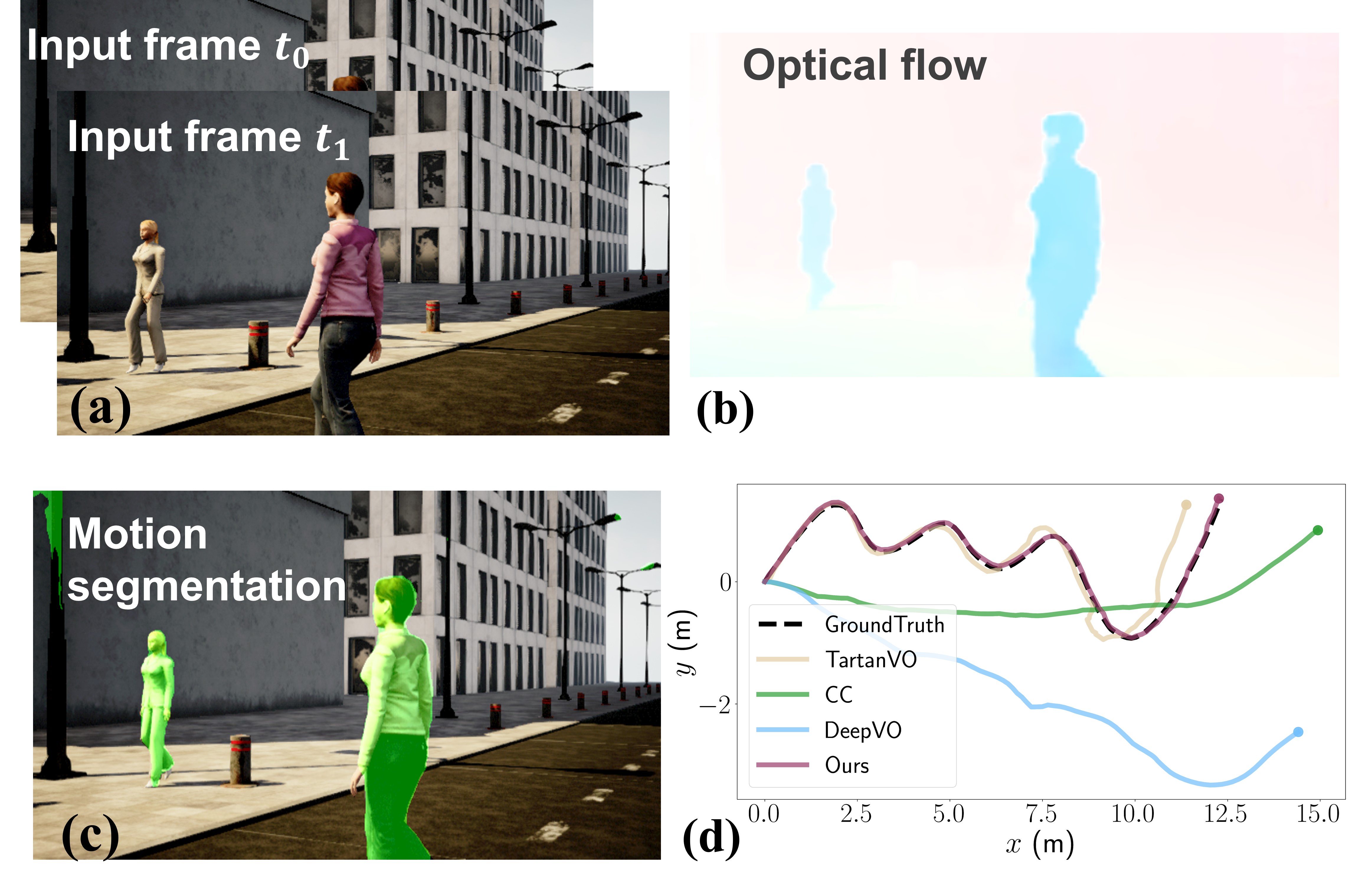}
    \caption{A overview of the DytanVO. (a) Input frame at time $t_0$ and $t_1$. (b) Optical flow output from the matching network. (c) Motion segmentation output after iterations. (d) Trajectory estimation on sequence RoadCrossing \uppercase\expandafter{\romannumeral6} from the AirDOS-Shibuya Dataset, which is a highly dynamic environment cluttered with humans. Ours is the only learning-based VO that keeps track. }
    \label{fig:cover_main}
    \vspace{-3mm}
\end{figure}

\vspace{-4mm}
To identify dynamic objects, object detection or semantic segmentation techniques are largely relied on to mask all movable objects, such as pedestrians and vehicles~\cite{liu2019visual, xu2019mid,li2017rgb, sun2017improving}. Their associated features are discarded before applying geometry-based methods. However, there are two issues of utilizing semantic information in dynamic VO. First, class-specific detectors for semantic segmentation heavily depend on appearance cues but not every object that can move is present in the training categories, leading to false negatives. Second, even if all moving objects in a scene within the categories, algorithms could not distinguish between ``actually moving" versus ``static but being able to move". In dynamic VO where static features are crucial to robust ego-motion estimation, one should segment objects based on pure motion (motion segmentation) rather than heuristic appearance cues. 



Motion segmentation utilizes relative motion between consecutive frames to remove the effect of camera movement from the 2D motion fields and calculates residual optical flow to account for moving regions. But paradoxically, ego-motion cannot be correctly estimated in dynamic scenes without a robust segmentation. There exists such a mutual dependence between motion segmentation and ego-motion estimation that has never been explored in supervised learning methods. Therefore, motivated by jointly refining the VO and motion segmentation, we propose our learning-based dynamic VO (DytanVO). To our best knowledge, our work is the first supervised learning-based VO for dynamic environments. The main contributions of this paper are threefold:


\begin{itemize}
    \item A novel learning-based VO is introduced to leverage the interdependence among camera ego-motion, optical flow and motion segmentation.
    \item We introduce an iterative framework where both ego-motion estimation and motion segmentation can converge quickly within time constraints for real-time applications. 
    \item Among learning-based VO solutions, our method achieves state-of-the-art performance in real-world dynamic scenes without finetuning. Furthermore, our method performs even comparably with visual SLAM solutions that optimize trajectories on the backend.  
\end{itemize}

\begin{figure*}[t!]
    \centering
    \includegraphics[width = \textwidth]{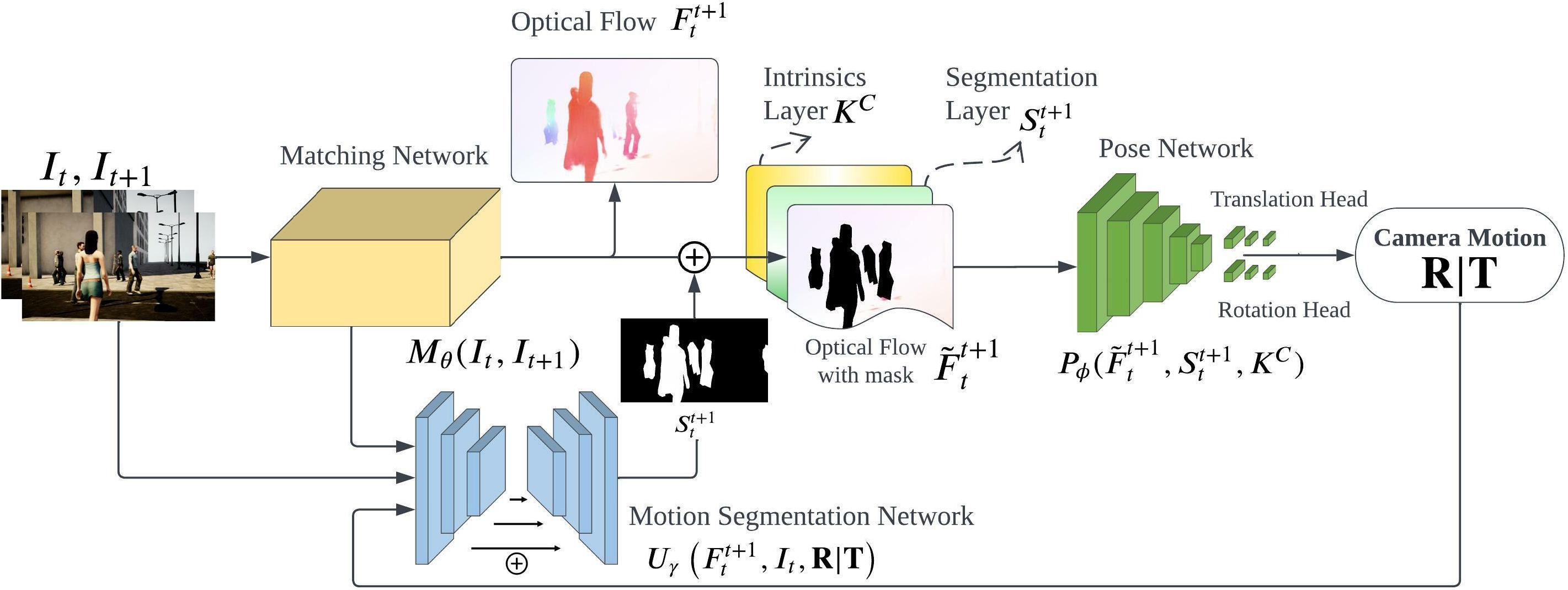}
    \caption{Overview of our three-stage network architecture. It consists of a matching network which estimates optical flow from two consecutive images, a pose network that estimates pose based on optical flow without dynamic movements, and a motion segmentation network that outputs a probability mask of the dynamicness. The matching network is forwarded only once while the pose network and the segmentation network are iterated to jointly refine pose estimate and motion segmentation. In the first iteration, we randomly initialize the segmentation mask. In each iteration, optical flow is set to zero inside masked regions. }
    \label{fig:architect}
    \vspace{-3mm}
\end{figure*}

\section{Related Work}
Learning-based VO solutions aim to avoid hard-coded modules that require significant engineering efforts for design and finetuning in classic pipelines~\cite{scaramuzza2011visual, fraundorfer2012visual}. For example, Valada ~\cite{valada2018deep} applies auxiliary learning to leverage relative pose information to constrain the search space and produce consistent motion estimation. 
Another class of learning-based methods rely on dense optical flow to estimate pose as it provides more robust and redundant modalities for feature association in VO~\cite{costante, tartanvo, zhan2020visual}. However, their frameworks are built on the assumption of photometric consistency which only holds in a static environment without independently moving objects. They easily fail when dynamic objects unpredictably cause occlusions or illuminations change. 

Semantic information is largely used by earlier works in VO or visual SLAM to handle dynamic objects in the scene, which is obtained by either a feature-based method or a learning-based method. Feature-based methods utilize hand-designed features to recognize semantic entities~\cite{kim2016effective}. An exemplary system proposed by~\cite{pillai2015monocular} computes SIFT descriptors from monocular image sequences in order to recognize semantic objects. On the other hand, data-driven CNN-based semantic methods have been widely used to improve the performance, such as DS-SLAM~\cite{yu2018ds} and SemanticFusion\cite{ mccormac2017semanticfusion}. A few works on semantic VO/SLAM have fused the semantic information from recognition modules to enhance motion estimation and vice versa~\cite{an2017semantic,lianos2018vso}. However, all these methods are prone to limited semantic categories, which leads to false negatives when scaling to unusual real-world applications such as offroad driving or MAV, and requires continuous efforts in ground-truth labeling. 

Instead of utilizing appearance cues for segmentation, efforts are made to segment based on geometry cues. FlowFusion~\cite{flowfusion} iteratively refines its ego-motion estimation by computing residual optical flow. GeoNet~\cite{yin2018geonet} divides its system into two sub-tasks by separately predicting static scene structure and dynamic motions. However, both depend on geometric constraints arising from epipolar geometry and rigid transformations, which are vulnerable to motion ambiguities such as objects moving in the colinear direction relative to the camera being indistinguishable from the background given only ego-motion and optical flow. On the other hand, MaskVO~\cite{xuan2022maskvo} and SimVODIS++~\cite{kim2022simvodis++} approach the problem by learning to mask dynamic feature points in a self-supervised manner. CC~\cite{ranjan2019competitive} couples motion segmentation, flow, depth and camera motion models which are jointly solved in an unsupervised way. Nevertheless, these self-supervised or unsupervised methods are trained on self-driving vehicle data dominated by pure translational motions with little rotation, which makes them difficult to generalize to completely different data distributions such as handheld cameras or drones. Our work introduces a framework that jointly refines camera ego-motion and motion segmentation in an iterative way that is robust against motion ambiguities as well as generalizes to the open world.

\section{Methodology}

\subsection{Datasets}
Built on TartanVO~\cite{tartanvo}, our method remains its generalization capability while handling dynamic environments in multiple types of scenes, such as car, MAV, indoor and outdoor. Besides taking camera intrinsics as an extra layer into the network to adapt to various camera settings as explored in~\cite{tartanvo}, we train our model on large amounts of synthetic data with broad diversity, which is shown capable of facilitating easy adaptation to the real world~\cite{wang2020tartanair, tobin2017domain, tremblay2018training}. 

Our model is trained on both TartanAir~\cite{wang2020tartanair} and SceneFlow~\cite{mayer2016large}. The former contains more than 400,000 data frames with ground truth of optical flow and camera pose in static environments only. The latter provides 39,000 frames in highly dynamic environments with each trajectory having backward/forward passes, different objects and motion characteristics. Although SceneFlow does not provide ground truth of motion segmentations, we are able to recover it by taking use of its ground truth of disparity, optical flow and disparity change maps.


\subsection{Architecture}
Our network architecture is illustrated in Fig. \ref{fig:architect}, which is based on TartanVO. Our method takes in two consecutive undistorted images ${I_t, I_{t+1}}$ and outputs the relative camera motion $\delta_t^{t+1} = (\Rv \vert \Tv)$, where $\Tv \in \mathbb{\Rv}^3$ is the 3D translation and $\Rv \in SO(3)$ is the 3D rotation. Our framework consists of three sub-modules, a matching network, a motion segmentation network, and a pose network. We estimate dense optical flow $F_t^{t+1}$ with a matching network, $M_\theta\left(I_t, I_{t+1}\right)$, from two consecutive images. The network is built based on PWC-Net \cite{pwcnet}. The motion segmentation network $U_\gamma$, based on a lightweight U-Net \cite{ronneberger2015u}, takes in the relative camera motion output, $\Rv \vert \Tv$, optical flow from $M_\theta$, and the original input frames. It outputs a probability map, $z_t^{t+1}$, of every pixel belonging to a dynamic object or not, which is thresholded and turned into a binary segmentation mask, $S_t^{t+1}$. The optical flow is then stacked with the mask and the intrinsics layer $K^C$, followed by setting all optical flow inside the masked regions to zeros, i.e., $\tilde{F}_t^{t+1}$. The last is a pose network $P\phi$, with ResNet50 \cite{resnet50} as the backbone, which takes in the previous stack, and outputs camera motion. 

\subsection{Motion segmentation}
Earlier dynamic VO methods that use motion segmentation rely on purely geometric constraints arising from epipolar geometry and rigid transformations~\cite{flowfusion, liu2019visual} so that they can threshold residual optical flow which is designed to account for moving regions. However, they are prone to catastrophic failures under two cases: (1) points in 3D moving along epipolar lines cannot be identified from the background given only monocular cues; (2) pure geometry methods leave no tolerance to noisy optical flow and less accurate camera motion estimations, which in our framework is very likely to happen in the first few iterations. Therefore, following~\cite{yang2021learning}, to deal with the ambiguities above, we explicitly model cost maps as inputs into the segmentation network after upgrading the 2D optical flow to 3D through optical expansion~\cite{yang2020upgrading}, which estimates the relative depth based on the scale change of overlapping image patches. The cost maps are tailored to coplanar and colinear motion ambiguities that cause segmentation failures in geometry-based motion segmentation. More details can be found in \cite{yang2021learning}. 

\subsection{Iteratively refine camera motion} \label{jointrefine}
We provide an overview of our iterative framework in Algorithm~\ref{alg:iteralgo}. During inference, the matching network is forwarded only once while the pose network and the segmentation network are iterated to jointly refine ego-motion estimation and motion segmentation. In the first iteration, the segmentation mask is initialized randomly using~\cite{french2020milking}. The criterion to stop iteration is straightforward, which is the rotational and translational differences of $\Rv \vert \Tv$ between two iterations being smaller than prefixed thresholds $\epsilon$. Instead of having a fixed constant to threshold probability maps into segmentation masks, we predetermine a decaying parameter that empirically reduces the input threshold over time, in order to discourage inaccurate masks in earlier iterations while embracing refined masks in later ones.

\begin{algorithm}[H]
    \caption{Inference with Iterations} \label{alg:iteralgo}
    \begin{algorithmic}
        \State Given two consecutive frames $I_{t},\ I_{t+1}$ and intrinsics $K$
        \State Initialize iteration number: $i \gets 1$ 
        \State Initialize difference in output camera motions: $\delta_{R\vert T} \gets \infty$ \vskip 2pt
        \State $^{i}F_t^{t+1} \gets \text{OpticalFlow}(I_t,\ I_{t+1})$
        \While{$\delta_{R\vert T} \geq \text{stopping criterion}, \epsilon$} \vskip 2pt
            \If{$i\ \text{is}\ 0$}
                \State $^{i}S_t^{t+1} \gets \text{getCowmask}(I_t)$ 
            \Else
                \State $^{i}z_t^{t+1} \gets \text{MotionSegmentation}(^{i}F_t^{t+1},\ I_t,\ ^{i}\Rv\vert^{i}\Tv)$ \vskip 2pt
                \State $^{i}S_t^{t+1} \gets \text{mask}\ ^{i}z_t^{t+1} \geq z_{threshold}$ \vskip 2pt
            \EndIf 
            \State $^{i}\tilde{F}_t^{t+1} \gets \text{set}\ ^{i}F_t^{t+1}=0\ \text{for}\ ^{i}S_t^{t+1}==1$ \vskip 2pt
            \State $^{i}\Rv\vert^{i}\Tv \gets \text{PoseNetwork}(^{i}\tilde{F}_t^{t+1},\ ^{i}S_t^{t+1},\ K)$ \vskip 2pt
            \State $\delta_{\Rv\vert\Tv} \gets\ ^{i}\Rv \vert^{i}\Tv \ -\ ^{i-1}\Rv\vert^{i-1}\Tv$ \vskip 2pt
            \State $i \gets i + 1$
        \EndWhile
    \end{algorithmic}
\end{algorithm}

Intuitively, during early iterations, the estimated motion is less accurate, which leads to false positives in the segmentation output (assigning high probabilities to static areas). However, due to the fact that optical flow map still provides enough correspondences regardless of cutting out non-dynamic regions from it, $P_\phi$ is able to robustly leverage the segmentation mask $S_t^{t+1}$ concatenated with $\tilde{F}_t^{t+1}$, and outputs reasonable camera motion. In later iterations, $U_\gamma$ is expected to output increasingly precise probability maps such that static regions in the optical flow map are no longer ``wasted" and hence $P_\phi$ can be improved accordingly.

\begin{figure}[t]
    \vspace{2mm}
    \centering
    \includegraphics[width = 0.88\columnwidth]{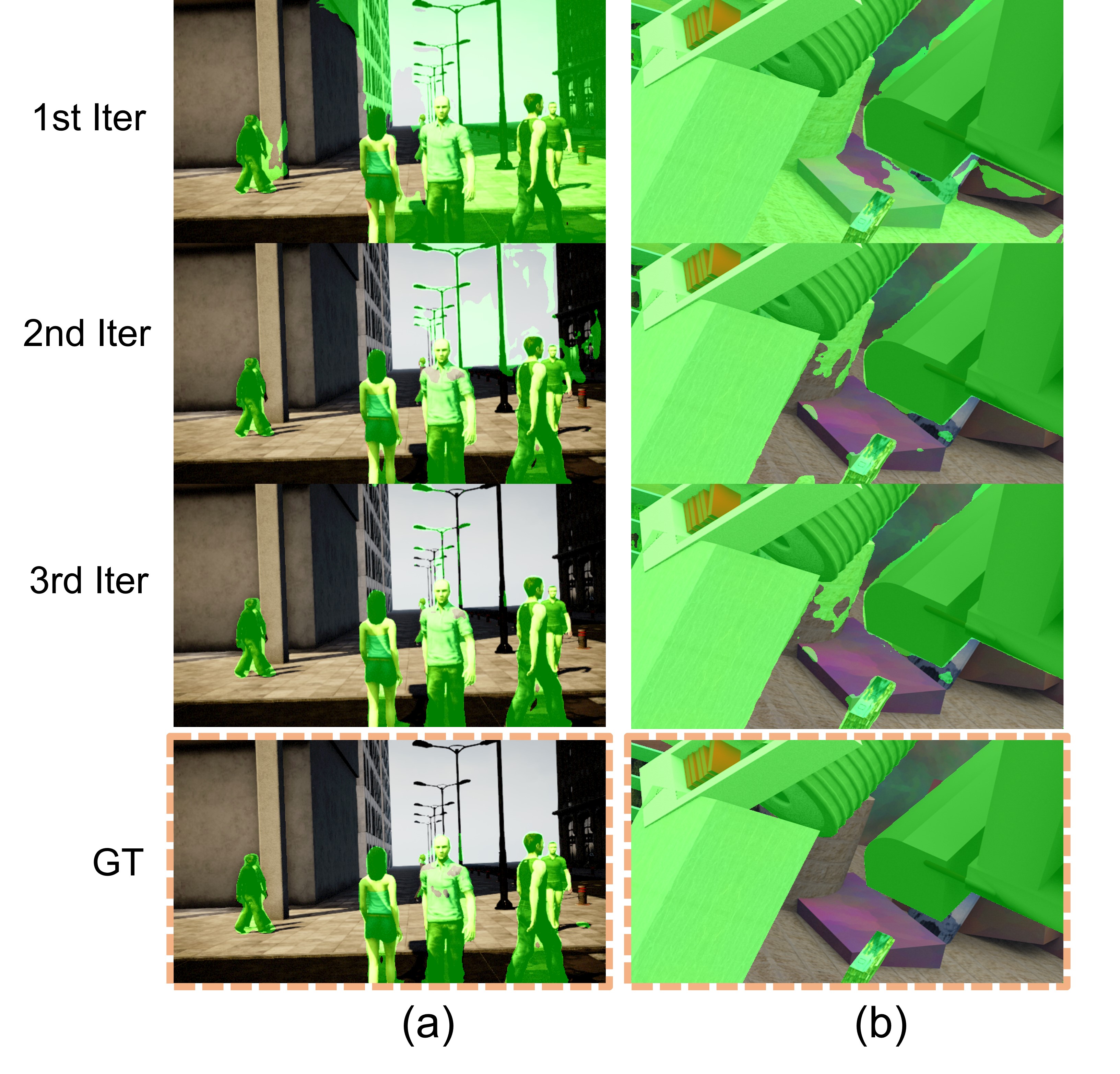}
    \caption{Motion segmentation output at each iteration when testing on unseen data. (a) Running inference on the hardest sequence in AirDOS-Shibuya with multiple people moving in different directions with our segmentation network. (b) Inference on the sequence from FlyingThings3D where dynamic objects take up more than 60\% area. Ground truth (GT) mask on Shibuya is generated by the segmentation network with GT ego-motion as input. }
    \label{fig:itermasks}
    \vspace{-5mm}
\end{figure}

In practice, we find that 3 iterations are more than enough to get both camera motion and segmentation refined. To clear up any ambiguity, a 1-iteration pass is composed of one $M_\theta$ forward pass and one $P_\phi$ forward pass with random mask, while a 3-iteration pass consists of one $M_\theta$ forward pass, two $U_\gamma$ forward passes and three $P_\phi$ forward passes. In Fig. \ref{fig:itermasks} we illustrate how segmentation masks evolve after three iterations on unseen data. The mask at the first iteration contains a significant amount of false positives but quickly converges beyond the second iteration. This verifies our assumption that the pose network is robust against false positives in segmentation results.

\subsection{Supervision}
We train our pose network to be robust against large areas of false positives. On training data without any dynamic object, we adopt the cow-mask \cite{french2020milking} to create sufficiently random yet locally connected segmentation patterns as a motion segmentation could occur in any size, any shape and at any position in an image while exhibiting locally explainable structures corresponding to the types of moving objects. In addition, we apply curriculum learning to the pose network where we gradually increase the maximum percentage of dynamic areas in SceneFlow from 15\%, 20\%, 30\%, 50\% to 100\%. Since TartanAir only contains static scenes, we adjust the size of the cow-masks accordingly. 

We supervise our network on the camera motion loss $L_P$. Under the monocular setting, we only recover an up-to-scale camera motion. We follow~\cite{tartanvo} and normalize the translation vector before calculating the distance to ground truth. Given ground truth motion $R \vert T$,
\begin{equation} \label{eq:lp}
    L_P=\left\| \frac{\hat{\Tv}}{\max(\Vert \hat{\Tv} \Vert, \epsilon )} - \frac{\Tv}{\max\left(\Vert \Tv \Vert, \epsilon \right)}  \right\| + \left\| \hat{\Rv} - \Rv \right\|
\end{equation}
where $\epsilon$=1$e$-6 to prevent numerical instability and $\hat{\cdot}$ denotes estimated quantities. 

Our framework can also be trained in an end-to-end fashion, in which case the objective becomes an aggregated loss of the optical flow loss $L_M$, the camera motion loss $L_P$ and the motion segmentation loss $L_U$, where $L_M$ is the L1 norm between the predicted flow and the ground truth flow whereas $L_U$ is the binary cross entropy loss between predicted probability and the segmentation label. 
\begin{equation} 
    L = \lambda_1 L_M + \lambda_2 L_U + L_P
    \vspace{-1mm}
\end{equation}
From preliminary empirical comparison, end-to-end training gives similar performance to training the pose network only, because we use $\lambda_1$ and $\lambda_2$ to regularize the objective such that the training is biased toward mainly improving the odometry rather than optimizing the other two tasks. This is ideal since the pose network is very tolerant of false positives in segmentation results (shown in \ref{jointrefine}). In the following section, we show our results of supervising only on Eq. \ref{eq:lp} by fixing the motion segmentation network.

\section{Experimental Results} \label{result}


\begin{table*}[tb]
\vspace{1mm}
\centering
\caption{ATE (m) results on dynamic sequences from AirDOS-Shibuya. Our method gives outstanding performance among VO methods. DeepVO, TrianFlow and CC are trained on KITTI only and unable to generalize to complex motion patterns. All SLAM methods use bundle adjustment (BA) on multiple frames to optimize the trajectory and hence we only numerically compare ours with pure VO methods. The best and the second best VO performances are highlighted as bold and underlined. We use ``-" to denote SLAM methods that fail to initialize. }
\label{tab:ate_shibuya}
\resizebox{0.9\textwidth}{!}{%
\begin{tabular}{clccccccc}
\toprule
\multicolumn{1}{l}{}         &                  & \multicolumn{2}{c}{\textbf{StandingHuman}} & \multicolumn{3}{c}{\textbf{RoadCrossing} (Easy)} & \multicolumn{2}{c}{\textbf{RoadCrossing} (Hard)} \\
\multicolumn{1}{l}{}         &                  & \uppercase\expandafter{\romannumeral1}                & \uppercase\expandafter{\romannumeral2}               & \uppercase\expandafter{\romannumeral3}            & \uppercase\expandafter{\romannumeral4}           & \uppercase\expandafter{\romannumeral5}           & \uppercase\expandafter{\romannumeral6}                   & \uppercase\expandafter{\romannumeral7}                  \\ \midrule
\multirow{5}{*}{\textbf{SLAM method}}  & DROID-SLAM~\cite{teed2021droid}       & 0.0051           & 0.0073          & 0.0103          & 0.0120          & 0.2778          & 0.0253              & 0.5788             \\
                              & AirDOS w/ mask~\cite{qiu2022airdos}   & 0.0606           & 0.0193          & 0.0951          & 0.0331          & 0.0206          & 0.2230              & 0.5625             \\
                              & ORB-SLAM w/ mask~\cite{orbslam} & 0.0788           & 0.0060          & 0.0657          & 0.0196          & 0.0148          & 1.0984              & 0.8476             \\
                              & VDO-SLAM~\cite{zhang2020vdo}         & 0.0994           & 0.6129          & 0.3813          & 0.3879          & 0.2175          & 0.2400              & 0.6628             \\
                              & DynaSLAM~\cite{bescos2018dynaslam}         & -                & 0.8836          & 0.3907          & 0.4196          & 0.4925          & 0.6446              & 0.6539             \\ \midrule
\multirow{5}{*}{\textbf{VO method}}    & DeepVO~\cite{wang2017deepvo}           & 0.3956           & 0.6351          & 0.7788          & 0.3436          & 0.5434          & 0.7223              & 0.9633             \\
                              & TrianFlow~\cite{trianflow}        & 0.9743           & 1.3835          & 1.3348               & 1.6172          & 1.4769          & 1.7154              & 1.9075             \\
                              & CC~\cite{ranjan2019competitive}               & 0.4527           & 0.7714          & 0.5406          & 0.6345          & 0.5411          & 0.8558              & 1.0896             \\
                              & TartanVO~\cite{tartanvo}         & {\underline {0.0600}}     & {\underline {0.1605}}    & {\underline {0.2762}}    & {\underline {0.1814}}    & {\underline {0.2174}}    & {\underline{0.3228}}        & {\underline {0.5009}}      \\ 
                               & Ours & \textbf{0.0327}  & \textbf{0.1017} & \textbf{0.0608} & \textbf{0.0516} & \textbf{0.0755} & \textbf{0.0365}     & \textbf{0.0660}   \\ \bottomrule
\end{tabular}%
}
\vspace{-3mm}
\end{table*}

\subsection{Implementation details}  \label{dataimp}

\subsubsection{Network} We intialize the matching network $M_\theta$ with the pre-trained model from TartanVO~\cite{tartanvo}, and fix the motion segmentation network $U_\gamma$ with the pre-trained weights from Yang et al. ~\cite{yang2021learning}. The pose network $P_\phi$ uses ResNet50~\cite{resnet50} as the backbone, removes the bach normalization layers, and adds two output heads for rotation $R$ and translation $T$. $M_\theta$ outputs optical flow at size of $H/4 \times W/4$. $P_\phi$ takes in a $5$-channel input, i.e., $\tilde{F}_t^{t+1}\in\mathbb{R}^{2\times H/4 \times W/4}$, $S_t^{t+1}\in\mathbb{R}^{H/4 \times W/4}$ and $K^C\in\mathbb{R}^{2\times H/4 \times W/4}$. The concatenation of $\tilde{F}_t^{t+1}$ and $K^C$ augments the optical flow input with 2D positional information while concatenating $\tilde{F}_t^{t+1}$ with $S_t^{t+1}$ encourages the network to learn dynamic representations.

\subsubsection{Training} Our method is implemented in PyTorch~\cite{paszke2017automatic} and trained on 2 NVIDIA A100 Tensor Core GPUs. We train the network in two stages on TartanAir, which includes only static scenes, and SceneFlow~\cite{mayer2016large}. In the first stage, we train $P_\phi$ independently using ground truth optical flow, camera motion, and motion segmentation mask in a curriculum-learning fashion. We generate random cow-masks~\cite{french2020milking} on TartanAir as motion segmentation input. Each curriculum is initialized with weights from the previous curriculum and takes 100,000 iterations with a batch size of 256. In the second stage, $P_\phi$ and $M_\theta$ are jointly optimized for another 100,000 iterations with a batch size of 64. During curriculum learning, the learning rate starts at 2$e$-4, while the second stage uses a learning rate of 2$e$-5. Both stages apply a decay rate of $0.2$ to the learning rate every 50,000 iterations. Random cropping and resizing (RCR)~\cite{tartanvo} as well as frame skipping are applied to both datasets. 

\subsubsection{Runtime} Although our method iterates multiple times to refine both segmentation and camera motion, we find in practice that 3 iterations are more than enough due to the robustness of $P_\phi$ as shown in Fig. \ref{fig:itermasks}. On an NVIDIA RTX 2080 GPU, inference takes 40ms with 1 iteration, 100ms with 2 iterations and 160ms with 3 iterations.

\subsubsection{Evaluation} We use the Absolute Trajectory Error (ATE) to evaluate our algorithm against other state-of-the-art methods including both VO and Visual SLAM. We evaluate our method on AirDOS-Shibuya dataset~\cite{qiu2022airdos} and KITTI Odometry dataset~\cite{geiger2013vision}. Additionally, in the supplemental material, we test our method on data collected in a cluttered intersection to demonstrate our method can scale to real-world dynamic scenes competitively.
\begin{table*}[t!]
\vspace{1mm}
\centering
\caption[ATE on Dynamic KITTI]{Results of ATE (m) on Dynamic Sequences from KITTI Odometry. Original sequences are trimmed into shorter ones that contain dynamic objects\protect\footnotemark. DeepVO~\cite{wang2017deepvo}, TrianFlow~\cite{trianflow} and CC~\cite{ranjan2019competitive} are trained on KITTI, while ours has not been finetuned on KITTI and is trained purely using synthetic data. Without backend optimization unlike SLAM, we achieve the best performance on 00, 02, 04, and competitive performance on the rest among all methods including SLAM. }
\label{tab:ate_dynakitti}
\resizebox{\textwidth}{!}{%
\begin{tabular}{clcccccccc}
\toprule
\multicolumn{1}{l}{\textbf{}} &                   & \textbf{00}                    & \textbf{01}             & \textbf{02}              & \textbf{03}              & \textbf{04}              & \textbf{07}              & \textbf{08}                        & \textbf{10}              \\ \hline
\multirow{4}{*}{\textbf{SLAM method}}  & DROID-SLAM~\cite{teed2021droid}        & 0.0148                & 49.193         & 0.1064          & 0.0119          & 0.0374          & 0.1939          & 0.9713              & 0.0368          \\
                                       & ORB-SLAM w/ mask~\cite{orbslam}  & 0.0187                & -              & 0.0796          & 0.1519          & 0.0198          & 0.2108          & 1.0479                    & 0.0246          \\
                                       & DynaSLAM~\cite{bescos2018dynaslam}          & 0.0138                & -              & 0.1046          & -               & 0.1450          & 0.3187          & 1.0559                     & 0,0264          \\ \midrule
\multirow{5}{*}{\textbf{VO method}}    & DeepVO~\cite{wang2017deepvo}            & (\underline{0.0206})                & 1.2896         & (0.2975)          & 0.0783          & 0.0506          & 0.7262          & (\textbf{0.6547})  & 0.1042          \\
                                       & TrianFlow~\cite{trianflow}         & 0.6966                & (8.2127)         & (1.8759)          & 1.6862          & 1.2950          & 1.5540               & (3.8984)                 & 0.2545          \\
                                       & CC~\cite{ranjan2019competitive}                & 0.0253  & (\textbf{0.3060}) & (0.2559)         & \underline{0.0505} & {\underline {0.0337}}    & {\underline{0.6789}}    & (1.0411)             & (\underline{0.0346})    \\
                                       & TartanVO~\cite{tartanvo}          & 0.0345                & 4.7080          & {\underline{0.1049}}    & 0.2832          & 0.0743          & 0.7108          & {\underline{0.9776}}              & 0.1024          \\  
                                       & Ours & \textbf{0.0126}       & {\underline{0.4081}}   & \textbf{0.0594} & \textbf{0.0406}          & \textbf{0.0180} & \textbf{0.6367} & 1.0344                   & \textbf{0.0280} \\ \bottomrule
\multicolumn{10}{l}{\small We use $(\cdot)$ to denote the sequence is in the training set of the corresponding method.}
\end{tabular}%
}
\end{table*}

\begin{figure*}[t!]
    \centering
    \vspace{2mm}
	\begin{subfigure}{0.245\textwidth}
	\includegraphics[width=1\textwidth]{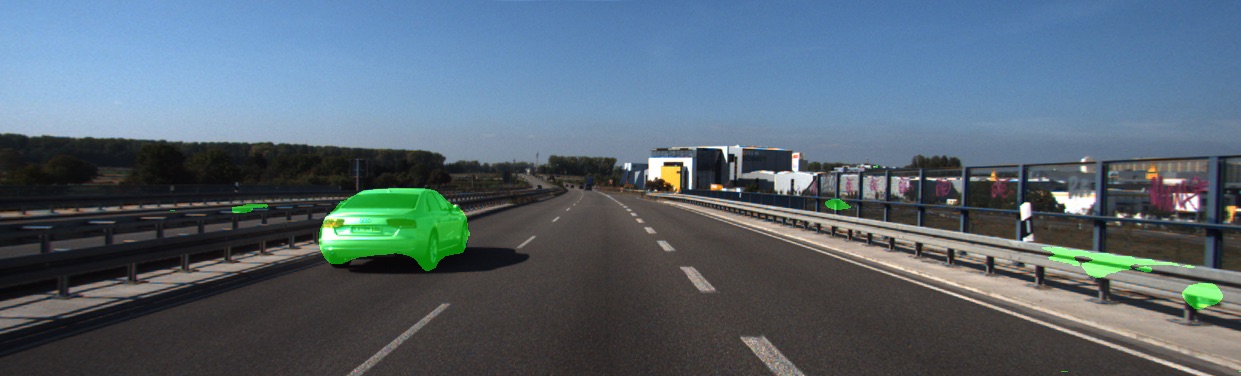}
	\end{subfigure}
	\begin{subfigure}{0.245\textwidth}
	\includegraphics[width=1\textwidth]{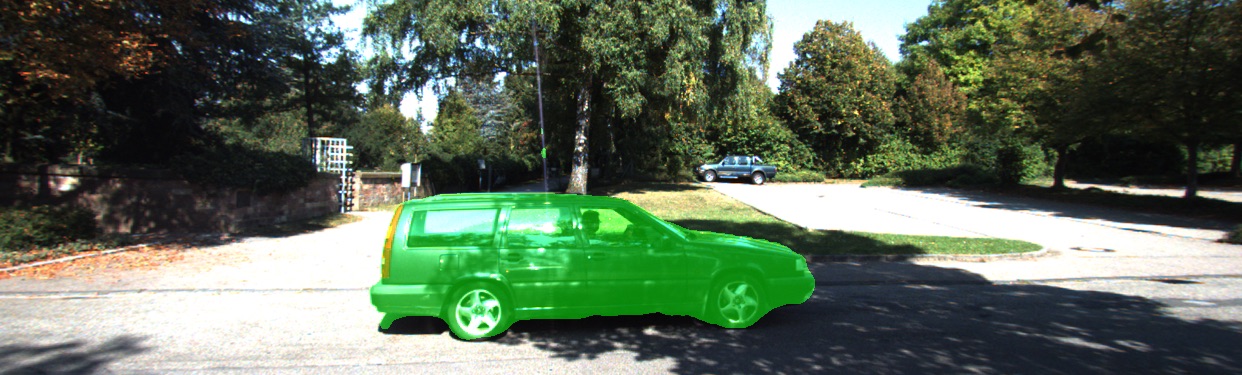}
	\end{subfigure} 
	\begin{subfigure}{0.245\textwidth}
	\includegraphics[width=1\textwidth]{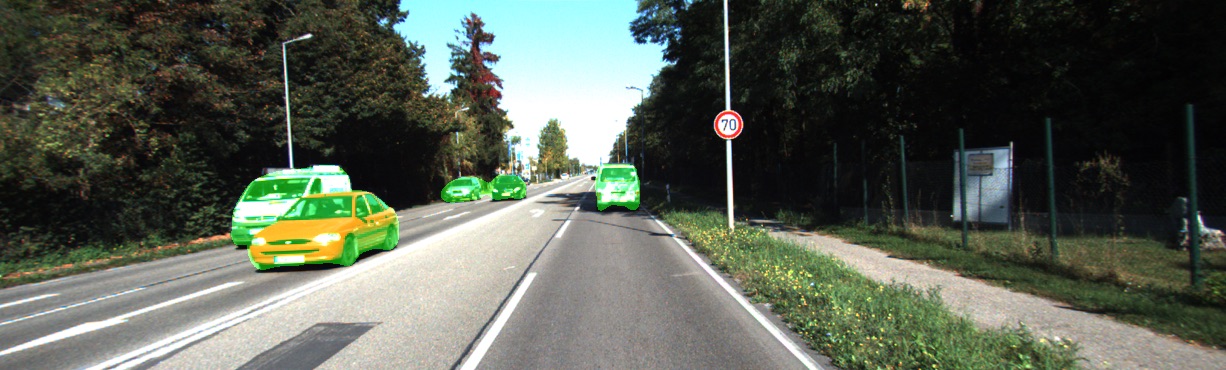}
	\end{subfigure} 
	\begin{subfigure}{0.245\textwidth}
	\includegraphics[width=1\textwidth]{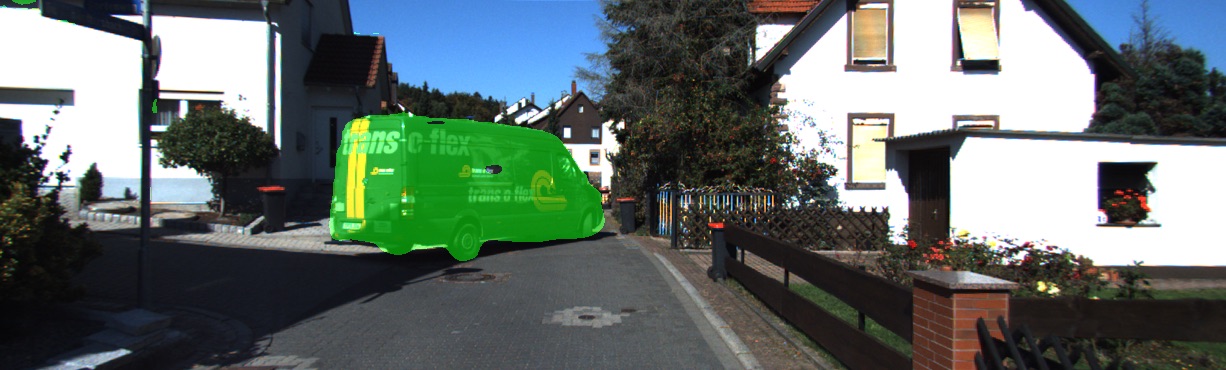}
	\end{subfigure}
	
	
	\begin{subfigure}{0.245\textwidth}
	\includegraphics[trim={0 2cm 0 0}, clip, width=1\textwidth]{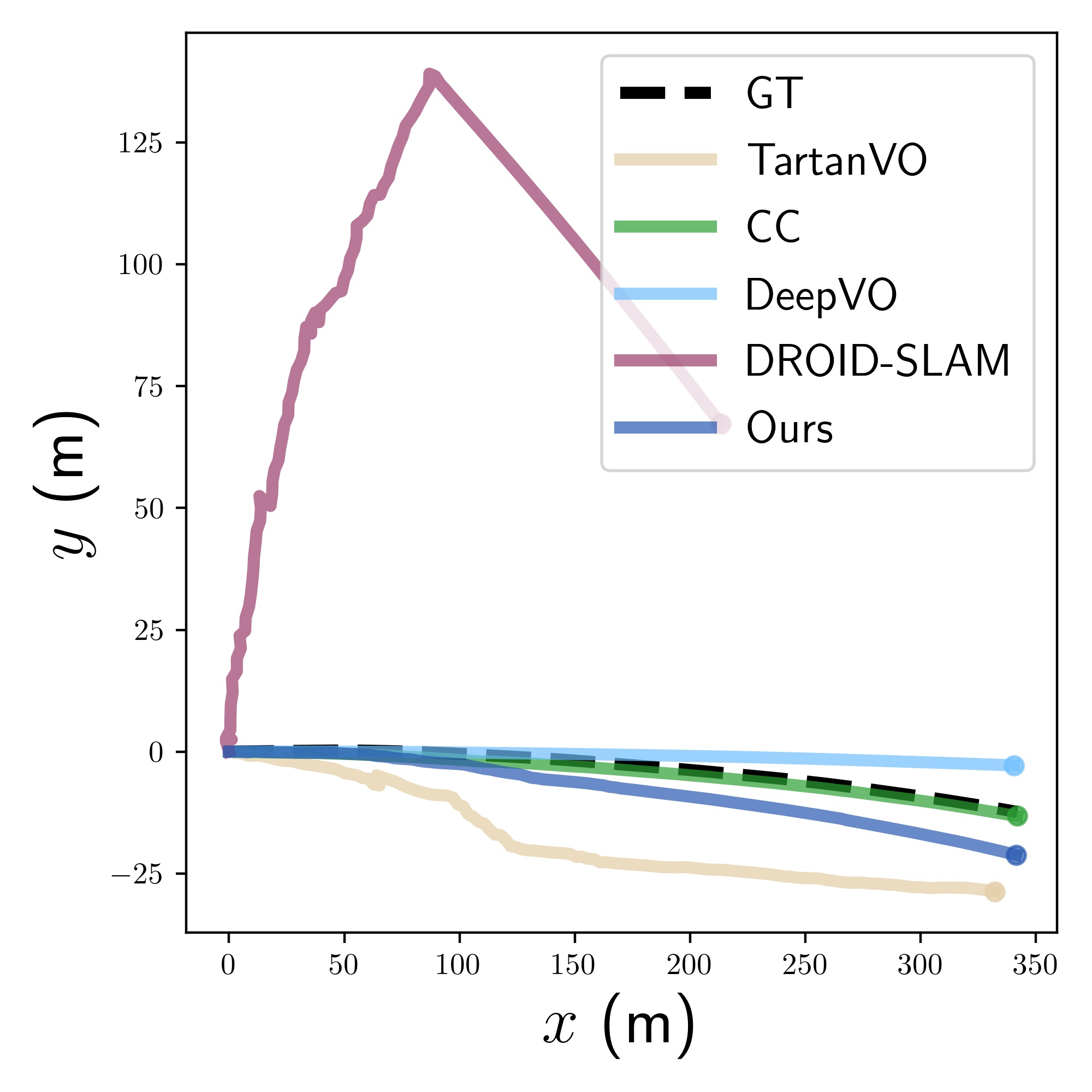}
	\end{subfigure}
	\begin{subfigure}{0.245\textwidth}
	\includegraphics[trim={0 2cm 0 0}, clip, width=1\textwidth]{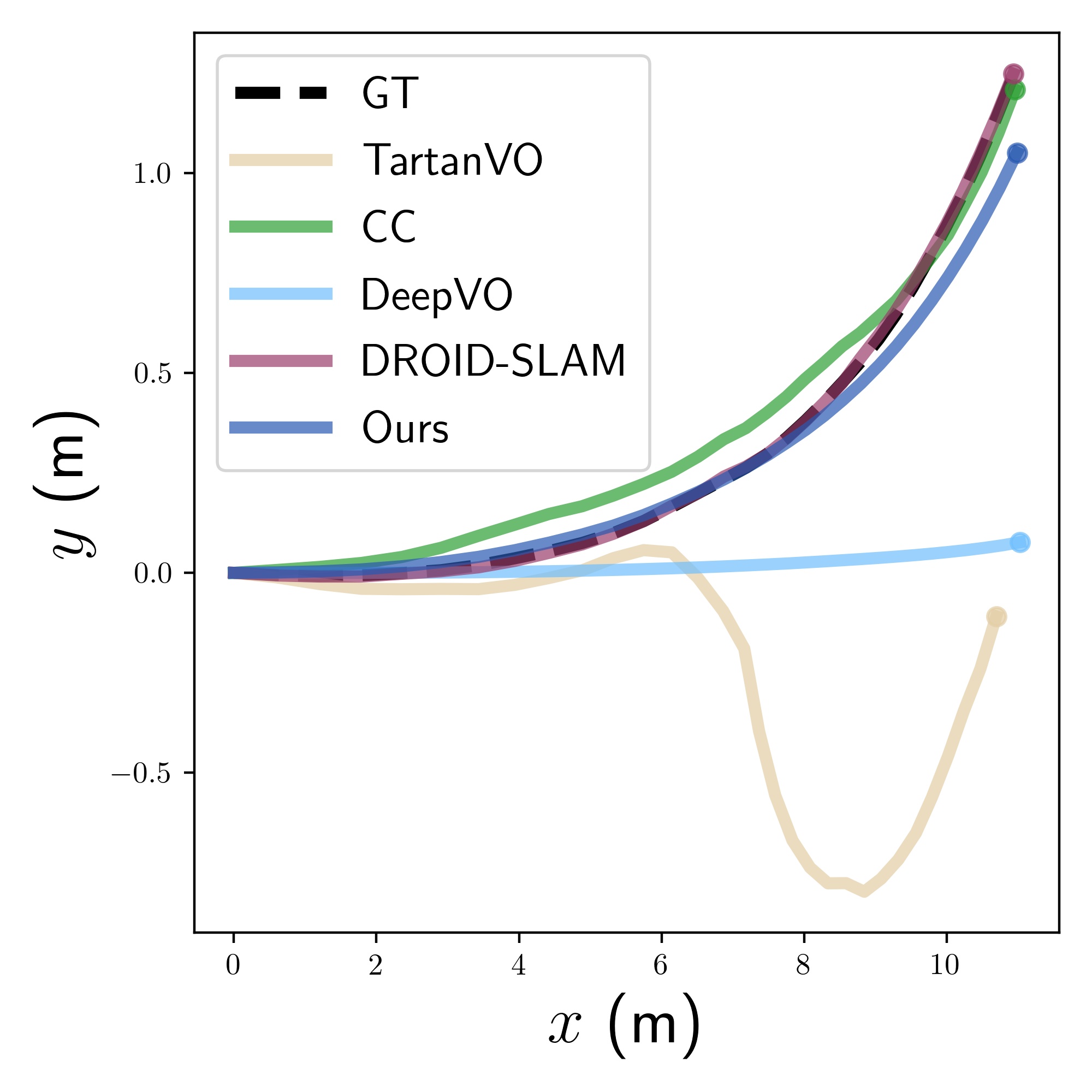}
	\end{subfigure} 
	\begin{subfigure}{0.245\textwidth}
	\includegraphics[trim={0 2cm 0 0}, clip, width=1\textwidth]{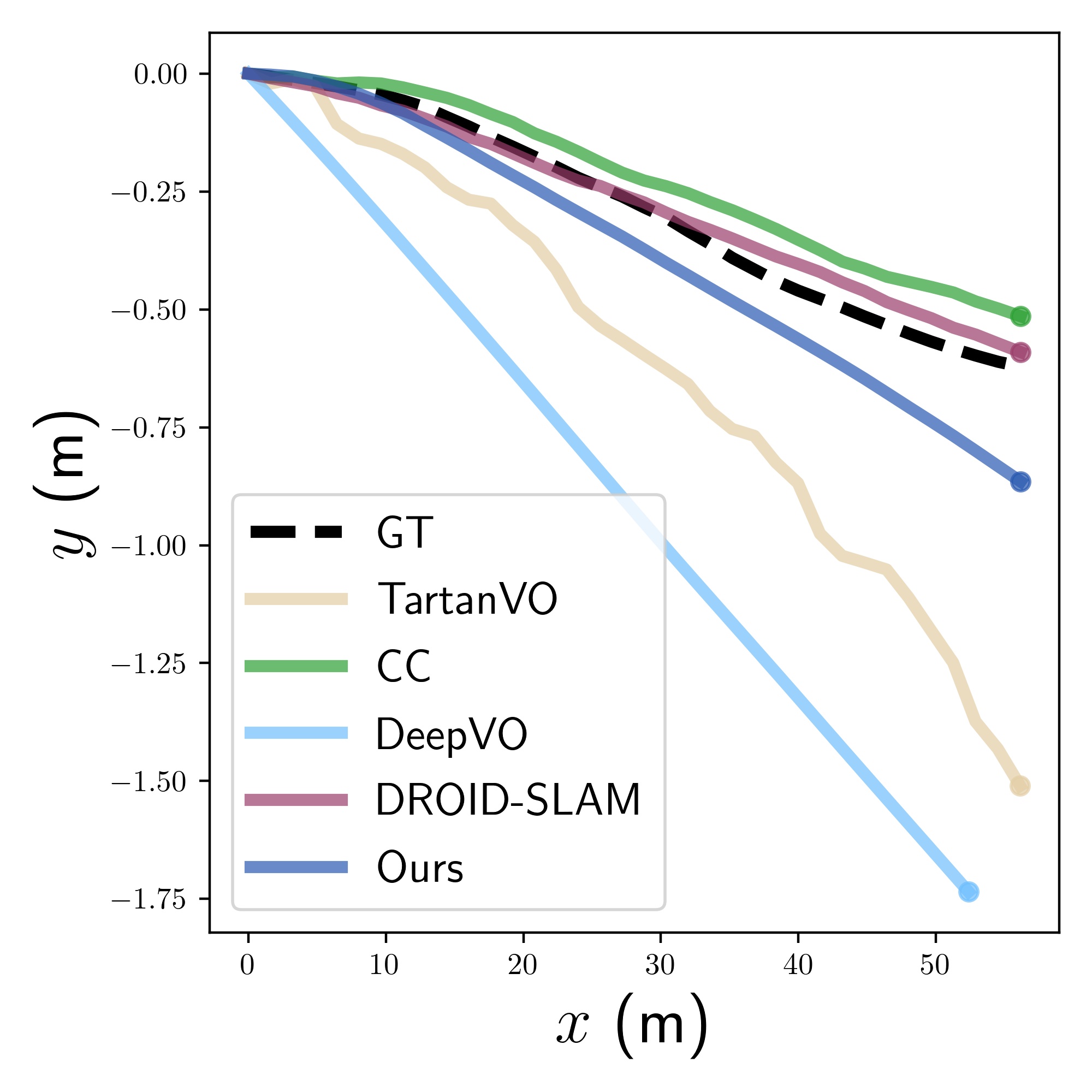}
	\end{subfigure} 
	\begin{subfigure}{0.245\textwidth}
	\includegraphics[trim={0 2cm 0 0}, clip, width=1\textwidth]{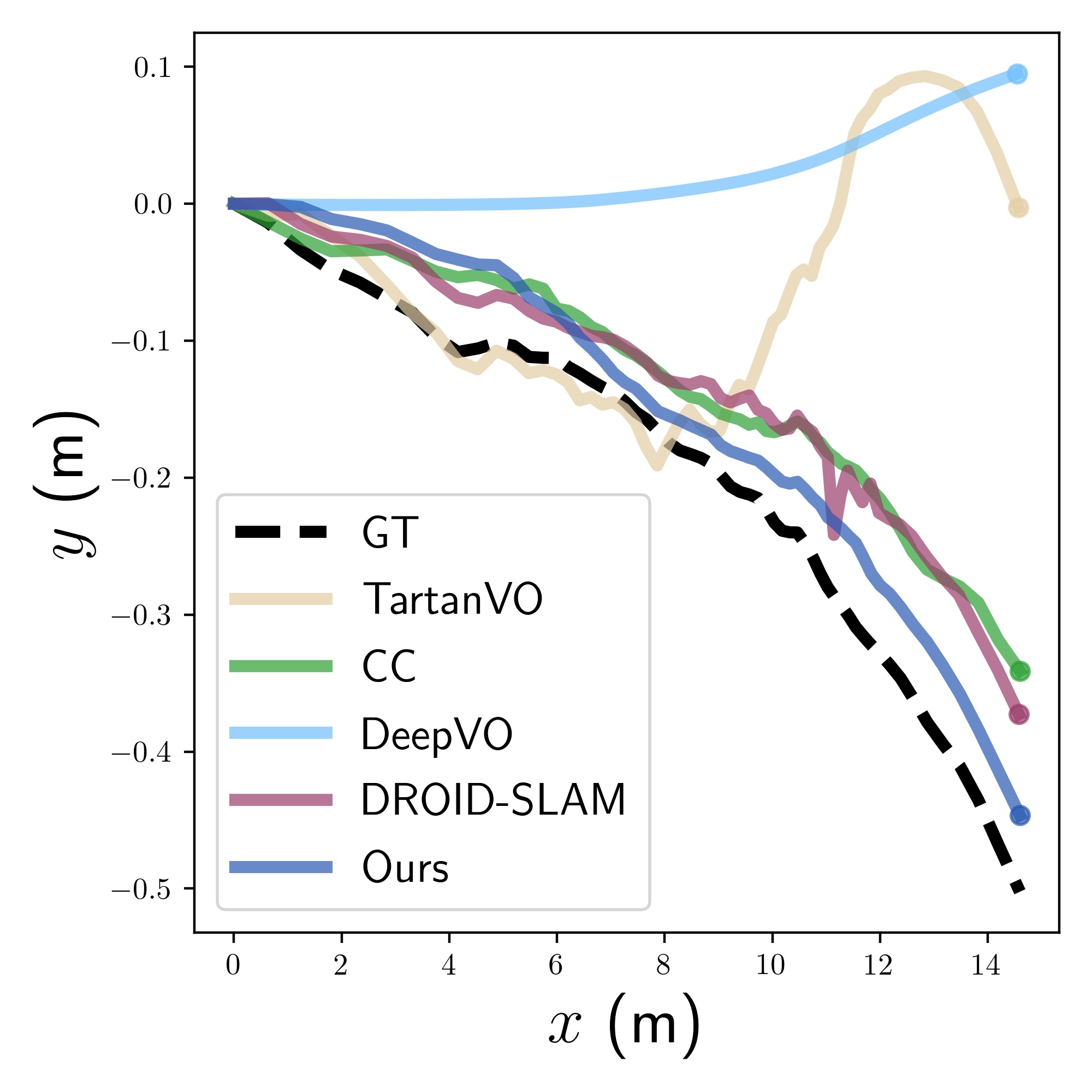}
	\end{subfigure}
    \caption{Qualitative results on dynamic sequences in KITTI Odometry 01, 03, 04 and 10. The first row is our segmentation outputs of moving objects. The second row is the visualization after aligning the scales of trajectories with ground truth all at once. Ours produces precise odometry given large areas in the image being dynamic even among methods that are trained on KITTI. Note that the trajectories do not always reflect the ATE results due to alignment.}
    \label{fig:kittitrajs}
    \vspace{-2mm}
\end{figure*}

\subsection{Performance on AirDOS-Shibuya Dataset}

We first provide an ablation study of the number of iterations (iter) in Tab. \ref{tab:ablation} using three sequences from AirDOS-Shibuya~\cite{qiu2022airdos}. The quantitative results are consistent with Fig. \ref{fig:itermasks} where the pose network quickly converges after the first iteration. We also compare the 3-iteration finetuned model after jointly optimizing $P_\phi$ and $M_\theta$ (second stage), which shows less improvement because the optical flow estimation on AirDOS-Shibuya already has high quality.

\begin{table}[H]
\centering
\caption{Experiments on number of iterations in ATE (m)}
\label{tab:ablation}
\resizebox{\columnwidth}{!}{%
\begin{tabular}{l|c|c|c}
\toprule
& {\textbf{Standing \uppercase\expandafter{\romannumeral1}}} & {\textbf{RoadCrossing \uppercase\expandafter{\romannumeral3}}} & {\textbf{RoadCrossing \uppercase\expandafter{\romannumeral7}}} \\ \midrule
\textbf{1 iter}  & 0.0649    & 0.1666     & 0.3157       \\
\textbf{2 iter}               & 0.0315    & 0.0974     & 0.0658       \\
\textbf{3 iter}               & 0.0327    & 0.0608     & 0.0660       \\
\textbf{Finetuned}   & 0.0384    & 0.0631     & 0.0531       \\ \bottomrule
\end{tabular}%
}
\vspace{-3mm}
\end{table}

We then compare our method with others on the seven sequences from AirDOS-Shibuya in Tab. \ref{tab:ate_shibuya} and demonstrate that our method outperforms existing state-of-the-art VO algorithms. This benchmark covers much more challenging viewpoints and diverse motion patterns for articulated objects than our training data. The seven sequences are categorized into three levels of difficulty: most humans stand still in Standing Human with few of them moving around, Road Crossing (Easy) contains multiple humans moving in and out of the camera's view, and in Road Crossing (Hard) humans enter camera's view abruptly. Besides VO methods, we also compare ours with SLAM methods that are able to handle dynamic scenes. DROID-SLAM~\cite{teed2021droid} is a learning-based SLAM trained on TartanAir. AirDOS~\cite{qiu2022airdos}, VDO-SLAM~\cite{zhang2020vdo} and DynaSLAM~\cite{bescos2018dynaslam} are three feature-based SLAM methods targeting dynamic scenes. We provide the performance of AirDOS and ORB-SLAM~\cite{orbslam} after masking the dynamic features during their ego-motion estimation. DeepVO~\cite{wang2017deepvo}, TartanVO and TrianFlow~\cite{trianflow} are three learning-based VO methods not targeting dynamic scenes while CC~\cite{ranjan2019competitive} is an unsupervised VO resolving dynamic scenes through motion segmentation. 


Our model achieves the best performance in all sequences among VO baselines and is competitive even among SLAM methods. DeepVO, TrianFlow and CC perform badly on AirDOS-Shibuya dataset because they are trained on KITTI only and not able to generalize. TartanVO performs better but it is still susceptible to the disturbance of dynamic objects. On RoadCrossing \uppercase\expandafter{\romannumeral5} as shown in Fig.~\ref{fig:cover_main}, all VO baselines fail except ours. In hard sequences where there are more aggressive camera movements and abundant moving objects, ours outperforms dynamic SLAM methods such as AirDOS, VDO-SLAM and DynaSLAM by more than 80\%. While DROID-SLAM remains competitive most time, it loses track of RoadCrossing \uppercase\expandafter{\romannumeral5} and \uppercase\expandafter{\romannumeral7} as soon as a walking person occupies a large area in the image. Note that ours only takes 0.16 seconds per inference with 3 iterations but DROID-SLAM takes extra 4.8 seconds to optimize the trajectory. More qualitative results are in the supplemental material.

\subsection{Performance on KITTI}

We also evaluated our method against others on sequences from KITTI Odometry dataset~\cite{geiger2013vision} in Tab.~\ref{tab:ate_dynakitti}. Our method outperforms other VO baselines in 6 out of 8 dynamic sequences with an improvement of 27.7\% on average against the second best method. DeepVO, TrianFlow and CC are trained on some of the sequences in KITTI while ours has not been finetuned on KITTI and is trained purely using synthetic data. Moreoever, we achieve the best ATE on 3 sequences among both VO and SLAM without any optimization. We provide qualitative results in Fig.~\ref{fig:kittitrajs} on four challenging sequences with fast-moving vehicles or dynamic objects occupying large areas in images. Note on sequence 01 which starts with a high-speed vehicle passing by, both ORB-SLAM and DynaSLAM fail to initialize, while DROID-SLAM loses track from the beginning. Even though CC uses 01 in its training set, ours gives only 0.1 higher ATE while 0.88 lower than the third best baseline. On sequence 10 when a huge van takes up significant areas in the center of the image, ours is the only VO that keeps track robustly. 

\footnotetext{Sequences listed are trimmed into lengths of $28, 133, 67, 31, 40, 136, 51$ and $59$ respectively which contain moving pedestrians, vehicles and cyclists.}

\subsection{Diagnostics}
While we observe our method is robust to heavily dynamic scenes with as much as 70\% dynamic objects in the image, it still fails when all foreground objects are moving, leaving textureless background only. This is most likely to happen when dynamic objects take up large areas in the image. For example, when testing on the test set of FlyingThings3D~\cite{mayer2016large} where 80\% of the image being dynamic, our method masks almost the entire optical flow map as zeros, leading to the divergence of motion estimation and segmentation. Future work could hence consider incorporating dynamic object-awareness into the framework and utilizing dynamic cues instead of fully discarding them. Additionally, learning-based VO tends to overfit on simple translational movements such as in KITTI, which is resolved in our method by training on datasets with broad diversity, but our method gives worse performance when there is little or zero camera motion, caused by the bias in currently available datasets. One should consider training on zero-motion inputs in addition frame skipping.

\section{Conclusion}

In this paper, we propose a learning-based dynamic VO (DytanVO) which can jointly refine the estimation of camera pose and segmentation of the dynamic objects. 
We demonstrate both ego-motion estimation and motion segmentation can converge quickly within time constrains for real-time applications.
We evaluate our method on KITTI Odometry and AirDOS-Shibuya datasets, and demonstrate state-of-the-art performance in dynamic environments without finetuning nor optimation on the backend. 
Our work introduces new directions for dynamic visual SLAM algorithms.

\section*{Acknowledgment}
This work was supported by CMU AirLab. We are grateful to Yuheng Qiu for running ORB-SLAM experiments as well as helping us with several revisions of the paper. Special thanks to Fei Yin and Zilin Zhang for the literature review. 


\bibliographystyle{ieeetr}
\bibliography{IEEEabrv,main}


\end{document}